\documentclass{acm_proc_article-sp}
\usepackage{hyperref}
\usepackage{graphicx}

\newtheorem{thm}{Theorem}

\newtheorem{lem}[thm]{Lemma}

\newcommand{\R}{\mathbb{R}}

\newcommand\norm[1]{|\hskip-.2ex|#1|\hskip-.2ex|}

\title{A Combinatorial Algorithm to Compute Regularization Paths}

\begin{document} 

\numberofauthors{3} 
%
\author{
%
%
\alignauthor
Bernd G\"artner\\
       \affaddr{ETH Zurich, Switzerland}\\
       \email{gaertner@inf.ethz.ch}
\alignauthor
Joachim Giesen\\
       \affaddr{Friedrich-Schiller-Universit\"at Jena, Germany}\\
       \email{giesen@informatik.uni-jena.de}
\and
\alignauthor
Martin Jaggi\\
       \affaddr{ETH Zurich, Switzerland}\\
       \email{jaggi@inf.ethz.ch}
\alignauthor
Torsten Welsch\\
       \affaddr{Friedrich-Schiller-Universit\"at Jena, Germany}
}
\date{27 March 2009}

\maketitle

\begin{abstract} 
For a wide variety of regularization methods, algorithms computing the entire solution path have been developed recently. Solution path algorithms do not only compute the solution for one particular value of the regularization parameter but the entire path of solutions, making the selection of an optimal parameter much easier. 
Most of the currently used algorithms are not robust in the sense that they cannot deal with general or degenerate input. Here we present a new robust, generic method for parametric quadratic programming. Our algorithm directly applies to nearly all machine learning applications, where so far every application required its own different algorithm.

We illustrate the usefulness of our method by applying it to a very low rank problem which could not be solved by existing path tracking methods, namely to compute part-worth values in choice based conjoint analysis, a popular technique from market research to estimate consumers preferences on a class of parameterized options.
\end{abstract} 

\\

\section{Introduction}
\label{sec:intro}

We study a combinatorial algorithm to solve parameterized quadratic
programs, i.e., to compute the whole solution path. Unlike other methods employed in machine learning, our algorithm can deal with singular objective function matrices, without perturbing the input. Regularization methods resulting in parametrized
quadratic programs have successfully been applied in many optimization, classification and regression tasks in a variety of areas as for example signal processing, statistics, biology, surface reconstruction and information retrieval. 
We will briefly review some applications here, and we will also study another application, namely choice based
conjoint analysis in more detail. Conjoint analysis comprises a
popular family of techniques mostly used in market research to assess
consumers' preferences on a set of options that are specified by
multiple parameters, see~\cite{GuHerHu00} for an overview and recent
developments. 
We will show that a regularization approach to the analysis of preference data
leads to a parameterized quadratic program with a sparse, low rank
positive semi-definite matrix describing the quadratic term of the
objective function. 

\subsection{Contributions and Related Work}

\paragraph{Solution Path Algorithms in Machine Learning}
An algorithm to compute the entire regularization path of the $C$-SVM has
originally been reported by Hastie et al.~\cite{Hastie:2004p3702}. \cite{Efron:2004p7599} gave such an algorithm for the LASSO, and later \cite{Loosli:2007p27} and \cite{GyeminLee:2007p4671} proposed solution path algorithms for $\nu$-SVM and one-class SVM respectively. Also Receiver Operating Characteristic (ROC) curves of SVM were solved by such methods \cite{Bach:2006p8823}. Support vector regression (SVR) is interesting as its underlying quadratic program depends on two parameters, a regularization parameter (for which the solution path was tracked by \cite{Gunter:2005p7622,Wang:2006p7363,Loosli:2007p27}) and a tube-width parameter (for which \cite{Wang:2008p7368} recently gave a solution path algorithm). See also \cite{Rosset:2007p6880} for a recent overview.

As Hastie et al.~\cite{Hastie:2004p3702} point out, one drawback of their algorithm for the two-class SVM is that it does not work for singular kernel matrices, but requires that in the process of the algorithm, all occurring principal minors of the kernel matrix need to be invertible. The same is required by the other existing path algorithms mentioned above. However, large kernel matrices do often have very low numerical rank, even in those cases where radial base function kernels are used \cite[Section 5.1]{Hastie:2004p3702}, but of course also in the case of linear SVMs with sparse features, such as in the application to conjoint analysis discussed in this paper. The inability to deal with singular sub-matrices is probably one of the main reasons that none of the above mentioned algorithms could so far effectively be applied on medium/larger scale problems~\cite{Hastie:2004p3702,Rosset:2007p6880}. \cite[Section 4.2]{Rosset:2007p6880} report that their algorithm prematurely terminates on $3\times3$ matrices due to this described problem.

By observing that all the above mentioned algorithms are reporting the solution paths of parametric quadratic programming of the form (\ref{eq:pQP}), 
we point out that it is in fact not necessary to use different algorithms for each problem variant. Generic algorithms have been known for quite some time
\cite{Murty:1988p6902,Ritter:1984p5860}, 
\cite{Benveniste:1981p5565,Best:1996p6334,Goldfarb:1999p6098}, 
\cite{Spjtvold:2008p6101}, but have interestingly not yet received
broader attention in the area of machine learning.

One goal of this paper is to popularize the generic solution algorithms for parametric quadratic programming,
because we think that they have some major advantages:
\vspace{-0.5\baselineskip}
\begin{itemize}
\item The same algorithm can be applied to any solution path problem that can be
 written in the form (\ref{eq:pQP}), which includes all of \cite{Hastie:2004p3702,Efron:2004p7599,Gunter:2005p7622,Wang:2006p7363,Bach:2006p8823,Loosli:2007p27,GyeminLee:2007p4671,Eigensatz:2008p8753,Wang:2008p7368}.
\item Many of the known generic algorithms can deal with \emph{all}
 inputs; in particular the algorithms can cope with singular sub-matrices in the objective function. 
\item There is significant existing literature on the performance,
 numerical stability, and complexity of the generic algorithms.
\item Our criss-cross algorithm is numerically more stable, and also more robust in the sense that small errors do not add up while tracking the solution path. Also, such algorithms are
 faster for sparse problems as in linear SVMs and conjoint analysis, because they do not need any matrix inversions.
\end{itemize}
\vspace{-1\baselineskip}

\paragraph{Comparison with other ways to deal with degeneracies}
Instead of using our described generic criss-cross method, another obvious way to avoid degeneracies caused by singular sub-matrices in the objective function is to add a small value $\varepsilon$ to each diagonal entry of the original matrix $Q$; subsequently, all simple methods for the regular case such as \cite{Hastie:2004p3702,Rosset:2007p6880} can be used. There are several problems with
this approach. First of all, the rank of the objective function
matrix is blown up artificially, and the potential of using efficient
small-rank-QP methods would be wasted. Secondly, the solution path of the
perturbed problem may differ substantially from that of the original
problem; in particular, the perturbation may lead to a much higher
number of bends and therefore higher tracking cost, and the computed
solutions could be far off the real solutions. In contrast, our criss
cross method avoids all these issues, since it always solves the
original unperturbed problem.

\section{Parametric Quadratic Programming}
\label{sec:pqp}

A quadratic program (QP) is the problem of minimizing a convex
quadratic function subject to linear equality and inequality
constraints. 
%
%
Here, we are interested in \emph{parameterized} quadratic programs
(pQPs) of the (standard) form
\begin{equation}
\label{eq:pQP}
\begin{array}{llll}
\mbox{\bf QP$(\mu)$} &\mbox{minimize}_x   & x^TQx + c(\mu)^Tx\\
        &\mbox{subject to} & Ax \geq b(\mu)\\
        &                  & x \geq 0,
\end{array}
\end{equation}
where $c:\R\rightarrow\R^n$ and $b:\R\rightarrow\R^m$ are functions
that describe how the linear term of the objective function and the
right-hand side of the constraints vary with some real parameter
$\mu$. $Q$ is an $n\times n$ symmetric positive semidefinite (PSD)
matrix (the quadratic quadratic term of the objective function), $c$
is an $n$-vector (the linear term of the objective function), $A$ is
an $m\times n$ matrix (the constraint matrix), and $b$ is an
$m$-vector (the right-hand side of the constraints).

Our goal is to solve a given problem \mbox{QP$(\mu)$} of the form
(\ref{eq:pQP}) for all $\mu$ in a given interval $[\mu_{\min},
\mu_{\max}]$, where we assume for now that \mbox{QP$(\mu)$} has an
optimal solution for all $\mu$ in that interval (the general case is
easy to handle as well, see the remark in the ``Odds and Ends''
paragraph below).
In other words, given any value of $\mu\in [\mu_{\min}, \mu_{\max}]$,
we want to retrieve an optimal solution $x^*$ to \mbox{QP$(\mu)$}
quickly, without having to solve the
problem from scratch. The task of solving such a problem for all
possible values of the parameter $\mu$ is called \emph{parametric
 quadratic programming}. What we want as output is a \emph{solution
 path}, an explicit function $x^*:\R\rightarrow\R^n$ that describes
the solution as a function of the parameter $\mu$.

It is well known that the solution path $x^*$ is piecewise linear if
$c$ and $b$ are linear functions of $\mu$, see for
example~\cite{Ritter:1962p6602}. 

\subsection{Regularization Methods and pQPs}

A variety of machine learning methods, in particular many regularization methods, are direct instances of parametric quadratic programming. Examples include support vector machines~\cite{Burges:1998p7916}, 
support vector regression~\cite{smola98SVR}
, the LASSO~\cite{Tibshirani:1996p5510}, surface reconstruction~\cite{implicitKernelSurface}, $\ell_1$-regularized least squares~\cite{SeungJeanKim:2007p5474}, and compressed sensing~\cite{Figueiredo:2007p5514}.

Let us shortly describe the support vector machine as a popular
example of a pQP that results from regularization. Later we will
re-discover the corresponding pQP in the context of choice based
conjoint analysis.


\paragraph{Support Vector Machine}

The support vector machine (SVM) is a standard tool for two-class
classification problems. In Section~\ref{sec:conjoint} we will see
that estimating part-worth values in choice based conjoint analysis can
be seen as a problem that is geometrically dual to binary
classification. The primal soft margin $C$-SVM is the following pQP:
\begin{equation}
\label{eq:svm}
\begin{array}{llll}
&\mbox{minimize}_{w,b,\xi} & 
\frac{1}{2}\norm{w}^2 + C \sum_{i=1}^n \xi_i\\
&\mbox{subject to} & y_i ( \omega^T x_i + b ) \ge 1 - \xi_i, \\
\end{array}
\end{equation}
where $y_i\in\{\pm 1\}$ is the class label of data point $x_i$ and $C$
is the regularization parameter. The dual of the soft margin $C$-SVM
is the following pQP (observe that the regularization parameter moves
from the objective function to the constraints):
\begin{equation}
\label{eq:dsvm}
\begin{array}{llll}
&\mbox{maximize}_\alpha   & \sum_{i} \alpha_i - \frac{1}{2}\sum_{i,j} \alpha_i \alpha_j y_i y_j x_i^T x_j \\
&\mbox{subject to} & \sum_{i} y_i \alpha_i = 1 \\
&                  & 0 \le \alpha_i \le C
\end{array}
\end{equation}

\section{The Criss-Cross Method for pQPs}
\label{sec:crisscross}

Next we will present a new generic algorithm that uses LCP techniques;
in contrast to Murty's method~\cite{Murty:1988p6902}, it uses the
extremely simple and elegant \emph{criss-cross method} as a
subroutine, resulting in what we believe is the simplest generic
algorithm that is able to deal with arbitrary PSD matrices $Q$.

The algorithm works in principle for more general continuous functions
$c$.  The main idea is to transform (\ref{eq:pQP}) to a parametric
\emph{linear complementarity problem} (LCP), and then use the
\emph{criss-cross method} to quickly update the solution while $\mu$
varies.

\subsection{The LCP Formulation}

Let us recall the Karush-Kuhn-Tucker optimality conditions for
quadratic programs, see e.g. \cite[Section 2.8]{lcp}.

\begin{thm}
An $n$-vector $x$ is an optimal solution to (\ref{eq:pQP}) if and
only if there exists and $n$-vector $u$ as well as $m$-vectors
$y$ and $v$ such that

\begin{tabular}{rrccl}
(i) & $v = Ax-b(\mu) \geq 0$ & and & $x\geq 0$ \\
(ii) & $u = c(\mu) - A^Ty + 2Qx \geq 0$ & and & $y\geq 0$ \\
(iii) & $x^Tu = 0$ & and & $y^Tv = 0$,
\end{tabular}
\end{thm}
where (i) encodes primal feasibility of $x$, (ii) encodes dual
feasibility of $y$, and (iii) is referred to as complementary
slackness.

The three conditions of the previous theorem can be rewritten in the
form
\begin{equation}
\label{eq:lcp}
\begin{array}{rcl}
w - Mz &=& q(\mu) \\
w, z &\geq& 0 \\
w^Tz &=& 0,
\end{array}
\end{equation}
where 
$w^T = (u^T, v^T)$, 
$z^T = (x^T,y^T)$, 
$q(\mu)^T = (c(\mu)^T,-b(\mu)^T)$ 
and  
$M = \left(\!\begin{array}{cc}2Q & -A^T \\ A & 0 \end{array}\!\right)$.

Problem (\ref{eq:lcp}) is a \emph{linear complementarity problem}
(LCP) with a PSD matrix (for all $w$, we have $w^TMw = 2u^TQu \geq 0$
--- this is why we have chosen the constraints to be ``$Ax\geq b$''
instead of the more common ``$Ax\leq b$''; the latter would lead to a
symmetric but not necessarily positive semidefinite matrix $M$ in the
LCP (\ref{eq:lcp})). In order so solve (\ref{eq:pQP}), we will therefore
find $w^*$ and $z^*$ that satisfy (\ref{eq:lcp}); then the first $n$
components of the $(n+m)$-vector $z^*$ form a solution to
(\ref{eq:pQP}). This reduction of QP to LCP is well-known, see
e.g. \cite[Section 1.2]{lcp}.

\subsection{The Criss-Cross Method} 

The criss-cross method is a combinatorial method for finding vectors
$w$ and $z$ that satisfy (\ref{eq:lcp}), given that $q=q(\mu)$ is
fixed (we address the case of varying $\mu$ below). The method is
guaranteed to terminate (with a solution, or a proof of infeasibility
of (\ref{eq:lcp})), given that $M$ is a \emph{sufficient} matrix, see
e.g.\ \cite{Fukuda:1998p6151}. This matrix class contains all PSD
matrices, meaning that the criss-cross method is applicable in our
setting. Our description below is for the special case of PSD matrices
\cite{Klafszky:1992p6245}.

The criss-cross method is an iterative method that goes through a
sequence of \emph{basic} solutions. To define such a solution, we
consider any subset $B\subseteq[k], k:=n+m$ and the matrix $M_B$ whose
$j$-th column is the $j$-th column $I_j$ of the $k\times k$ identity
matrix $I$ (if $j\in B$), or the $j$-th column of $-M$ (if $j\not\in
B$). $B$ is called a \emph{basis} if $M_B$ is invertible. For
example, $B=[k]$ is a basis since $M_{[k]}=I$.

Given a basis, we obtain the corresponding \emph{basic solution} as
the unique solution of the following system of equations:
\begin{eqnarray*}
z_j &=& 0, \quad j\in B, \\
w_j &=& 0, \quad j\not\in B, \\
w - Mz &=& q.
\end{eqnarray*}
This indeed has a unique solution, since substitution of the first two
sets of equations into $w-Mz=q$ yields the system $M_B \lambda_B = q$,
where $\lambda_j = w_j$ if $j\in B$ and $\lambda_j=z_j$ otherwise.

It is clear that every basic solution $(w,z)$ satisfies $w^Tz=0$, but
$w,z\geq 0$ may not hold. The criss-cross method tries to rectify this
by repeatedly moving to another basis and corresponding basic
solution, until $w,z\geq 0$ in which case the LCP is solved.

Given a basis $B$ along with $\lambda^*_B$ (the unique solution of
$M_B \lambda_B = q$), one step of the method works as follows. If
$\lambda^*:=\lambda^*_B\geq 0$, we are done; otherwise, choose the smallest
index $r$ 
such that $\lambda^*_r<0$. With respect to $B$, the system
$w-Mz=q$ can be written as $M_B\lambda_B + M_N\lambda_N = q$,
where $N=[k]\setminus B$. Consequently,
\[\lambda_B = M_B^{-1}q - M_B^{-1}M_N\lambda_N\]
for all solutions of $w-Mz=q$ (the basic solution associated with $B$
is obtained from $\lambda_N=0$).

Let the $k\times k$ matrix $\Lambda = -M_B^{-1}M_N$ be the
\emph{dictionary} associated with $B$, so that we have
\begin{equation}
\label{eq:dict}
\lambda_B = M_B^{-1}q + \Lambda\lambda_N.
\end{equation}
There are now two cases:
\begin{itemize}
\item[(a)] $\Lambda_{rj}\leq 0$ for all $j\in [k]$. By (\ref{eq:dict}) we
have
\[(\lambda_B)_r = \lambda^*_r + \Lambda^r \lambda_N\] for all
solutions of $w-Mz=q$, where $\Lambda^r$ is the $r$-th row of
$\Lambda$. But since this yields $\lambda_r<0$ whenever $\lambda_N\geq
0$, there can't be any solution to $w-Mz=q$ with $w,z\geq 0$, and we
can conclude that the LCP is infeasible.
\item[(b)] $\Lambda_{rj} > 0$ for some $j\in [k]$. Choose the smallest
index $s$ such that $\Lambda_{rs}>0$ and set $p:=\max(r,s)$. If
$\Lambda_{pp}\neq 0$, update $B$ to $B':=B\oplus\{p\}$ (\emph{diagonal pivot}),
otherwise update $B$ to $B':=B\oplus\{r,s\}$ (\emph{exchange pivot}),
where $\oplus$ denotes symmetric set difference.
\end{itemize}

\begin{lem}
The set $B'$ resulting from step (b) is again a basis.
\end{lem}
\begin{proof}
 In general, if $B'=B\oplus D$, then $M_{B'}$ is obtained from $M_B$
 by replacing the columns whose indices are in $D$ with the
 corresponding columns of $M_N$. This update can be written as
 \[M_{B'} = M_B T,\] where $T_j = I_j$ if $j\not\in D$ and $T_j =
 (M_B^{-1}M_N)_j = -\Lambda_j$ for $j\in D$. Moreover, since $M_B$
 was invertible, $M_{B'}$ is invertible if and only if $\det(T)\neq
 0$. If $D=\{p\}$ (the diagonal pivot), we get
 $\det(T)=-\Lambda_{pp}\neq 0$. If $D=\{r,s\}$ (the exchange pivot),
 we assume w.l.o.g.\ $r<s=p$ and get
 \[
 \det(T)= \det\left(\begin{array}{cc}
     \Lambda_{rr} & \Lambda_{rs}\\
     \Lambda_{sr} & 0
   \end{array}
 \right).
 \]
 In order to evaluate this, we need one observation concerning the
 structure of $\Lambda=-M_B^{-1}M_N$. Let us call an $n\times n$
 matrix \emph{bisymmetric} if it is of the form
 $\left(\begin{array}{cc}Q & -A^T \\ A & P\end{array}\right)$ where
 both $Q$ and $P$ are symmetric. For example,
 $M=-M_{[k]}^{-1}M_{\emptyset}$ is bisymmetric, but simple
 calculations show that $\Lambda=-M_{B}^{-1}M_N$ is also bisymmetric,
 hence $\Lambda_{sr}=-\Lambda_{rs}<0$ which implies $\det(T)>0$. 
\end{proof}

This method is due Klafszky and Terlaky \cite{Klafszky:1992p6245} who
also show that it terminates after having gone through a finite number
of bases.

\subsection{Varying the Parameter} 

We now turn to the case where the right-hand side $q(\mu)$ of
(\ref{eq:lcp}) varies. Assume that we have solved the problem for
$\mu=\mu_{\min}$ using the criss-cross method, meaning that we now
have a basis $B\subseteq[k]$ such that
\[\lambda^*_B(\mu) = M_{B}^{-1}q(\mu)\geq 0.\]
Since $\lambda^*_B(\mu)$ depends linearly on $\mu$ (assuming that $b$
and $c$ in (\ref{eq:pQP}) are linear functions), we can easily compute
the largest value $\mu'\geq\mu$ such that $\lambda^*_B(\mu')\geq 0$
(we may have $\mu'=\mu$ but also $\mu'=\infty$).

For every value $\kappa\in[\mu,\mu']$, $\lambda^*_B(\kappa)$ is still
a solution to (\ref{eq:lcp}) with right-hand side $q(\kappa)$. In
order to be able to trace the solution beyond $\kappa=\mu'$, we apply
the criss-cross method to (\ref{eq:lcp}) again, starting from the
basis $B$, but now with the right-hand side $q=q(\mu'+\varepsilon)$,
where $\varepsilon$ is a symbolic parameter meant to represent an
arbitrarily small positive value. That way, we solve a slightly
perturbed LCP, starting from a solution to the old LCP, and in
practice, we expect that this will take only very few
iterations. There are no theoretical guarantees for this,
though\footnote{This complexity behavior is expected to be very
 similar to running Simplex steps for a slightly perturbed linear
 program, starting from a solution for the original problem.}.

In running the criss-cross-method on the symbolically perturbed
problem, all values $\lambda^*_r$ whose signs are being used to check
whether we currently have a solution to (\ref{eq:lcp}) are linear
polynomials in $\varepsilon$ (dictionary entries that are needed to
check for infeasibility are unaffected by $\varepsilon$). The sign of
a linear polynomial $\sigma+\varepsilon \tau$ is determined by
$\sigma$ if $\sigma\neq 0$, and by $\tau$ otherwise.

It follows that for the basis $B'$ obtained upon
termination of the criss-cross method, there are $k$-vectors $s$ and $t$
such that
\begin{equation}
\label{eq:eps}
\lambda^*_{B'}(\mu'+\varepsilon) = s + \varepsilon t,
\end{equation}
where $s_j>0$ or $s_j=0, t_j \geq 0 \quad \forall j\in[k]$.

This implies that $\lambda^*_{B'}(\mu'+\varepsilon)\geq 0$ for any
sufficiently small \emph{numerical} value of $\varepsilon$. In other
words, $B'$ is valid throughout a whole interval $[\mu',\mu'+\varepsilon']$,
where $\varepsilon'>0$ is easy to compute from (\ref{eq:eps}).

While increasing $\mu$, we therefore subdivide our interval
$[\mu_{\min},\mu_{\max}]$ into pieces over which the solution to
(\ref{eq:lcp}) and therefore also the solution to (\ref{eq:pQP}) is
linear in $\mu$. There are only finitely many such pieces, since no
basis $B$ can repeat (if $B$ is valid for two values $\mu,\mu'$, it is
also valid for any intermediate value).

\paragraph{Performance} By the above analysis, we have that our
algorithm calculates the entire solution path of any parametric
quadratic program in finite time. Also, it is well suited to make use
of the sparseness of the solutions, which is a key property of all
regularization methods. When running the algorithm, the relevant size
of the matrices $M_B$ that we have to deal with is bounded by the
number the number of non-zero entries in $x$, plus $m$.

\paragraph{Odds and Ends} The solution path computed in the above way
may be discontinuous, since the solution to the LCP may ``jump'' when
we move from $q(\mu')$ to $q(\mu'+\varepsilon)$. This is due to the
fact that the LCP has in general not a unique solution, and the
criss-cross method has no control over which optimal solution it
finds. However if one strictly wants continuity, one can simply
insert connecting straight-line segments: Since both endpoints are
solutions for $q(\mu')$ (set $\varepsilon=0$), all intermediate points
will be solutions as well. This holds for the $x$-part of $(w,z)$ (the
QP solution) by convexity of the optimal region in (\ref{eq:pQP}), but it
also holds for $(w,z)$ w.r.t.\ the LCP by a result of Adler and Gale
\cite{Adler:1975p7102}.

For the above to work, we do not even have to assume that
\mbox{QP$(\mu)$} has an optimal solution throughout
$[\mu_{\min},\mu_{\max}]$. Our method can handle the general case. We
may start off at $\mu=\mu_{\min}$ with an unsolvable LCP (the
criss-cross method will report this), or we may run into an unsolvable
situation later. In order to trace $\mu$ through such a situation, we
simply choose the ``next event'' as the largest $\mu'\geq \mu$ for
which $(\lambda^*_{B})(\mu')_r \leq 0$, where $(\lambda_B)_r$ is the
variable for which infeasibility was detected in case (a) of the
criss-cross method.

\section{Choice Based Conjoint Analysis}
\label{sec:conjoint}

In general conjoint analysis includes two tasks: (a) preference data
assessment, and (b) analysis of the assessed data. In choice based
conjoint analysis (CBC) preference data are assessed on a set of
options $A$ in a sequence of choice experiments. In every choice
experiment a consumer has to choose the most preferred option out of a
few options that are presented to her/him (typically between two and
five options from $A$). The set of all options is assumed to carry a
conjoint structure, i.e., $A$ is the Cartesian product
$A=A_1\times\ldots\times A_n$ of parameter sets $A_i$. In the
following we assume that the parameter sets $A_i$ are finite. Choice
data are of the form $a \succeq b$, where $a=(a_1,\ldots,a_n),
b=(b_1,\ldots,b_n)\in A$ and $a$ was preferred over $b$ by some
consumer in a choice experiment. Our goal is to compute an interval
scale $v:A\rightarrow\mathbb{R}$ on the domain $A$ from a set of
choice data. The scale $v$ is meant to represent the preferences of
the population of consumers who contributed to the choice experiments,
i.e., $a\in A$ is more popular than $b\in A$ if $v(a)>v(b)$, and the
difference $v(a)-v(b)$ tells how much more popular $v(a)$ is than
$v(b)$.

In the data analysis stage of conjoint analysis it is almost always
assumed~\cite{sawtooth,ChapelleH04} that the scale $v$ is linear,
i.e., that it can be written as as
\begin{equation}
v(a) = v\big((a_1,\ldots,a_n)\big) = \sum_{i=1}^n v_i(a_i),
\end{equation}
where $v_i:A_i\rightarrow\mathbb{R}$ are also interval scales, see~\cite{Keeney93} for a justification of choosing a linear scale. The value
$v_i (a_i), a_i\in A_i$ is called the part-worth value of level
$a_i$, i.e., the value that it contributes to the overall value of
an option $a$ where the level $a_i$ is present. The goal of choice
based conjoint analysis is to compute/estimate part-worth values for
all attribute levels from choice data.

\paragraph{Regularization approach to compute part-worth values}

\begin{figure}
\begin{center}
 \includegraphics[width=0.15\textwidth]{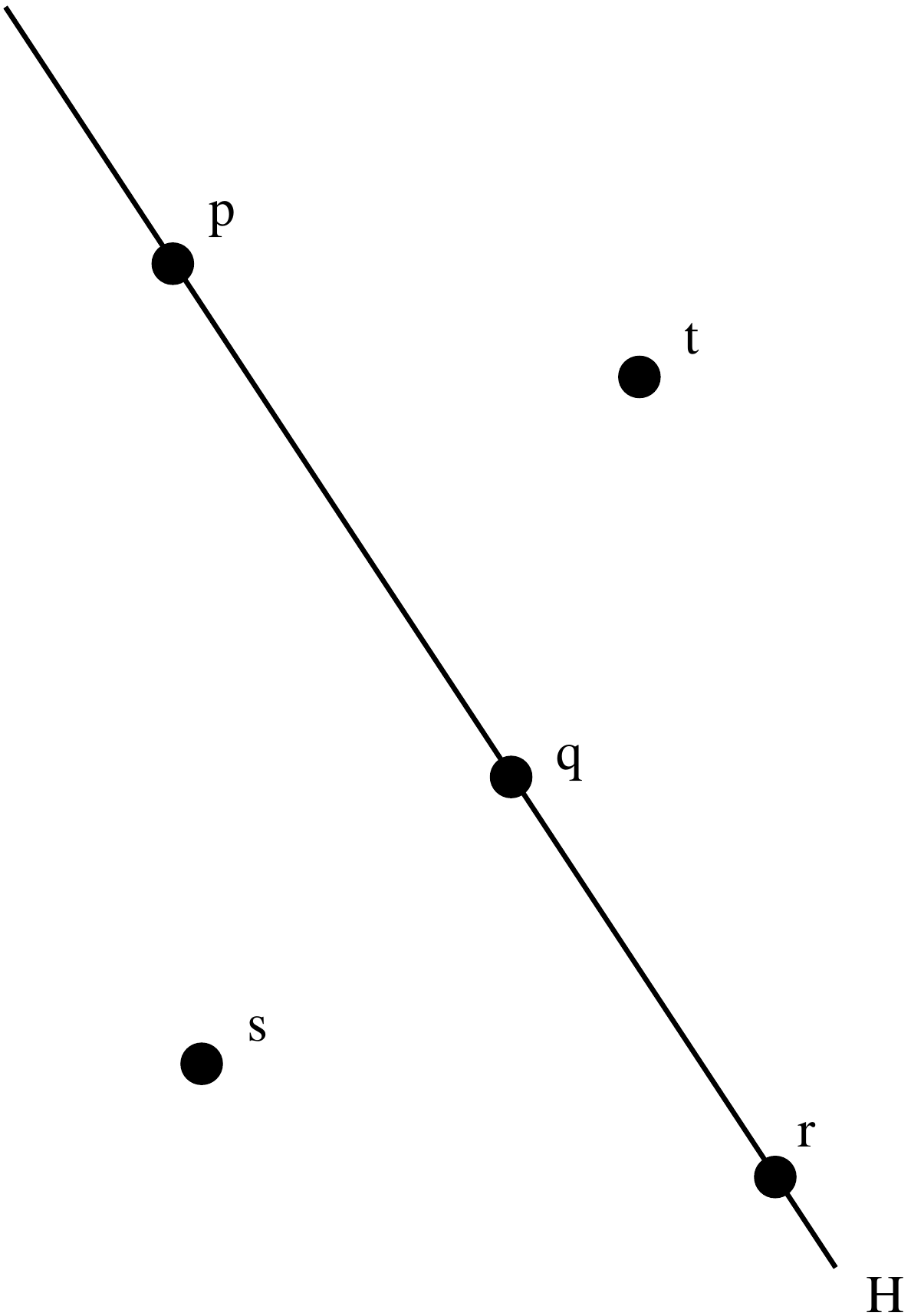} \qquad
 \includegraphics[width=0.22\textwidth]{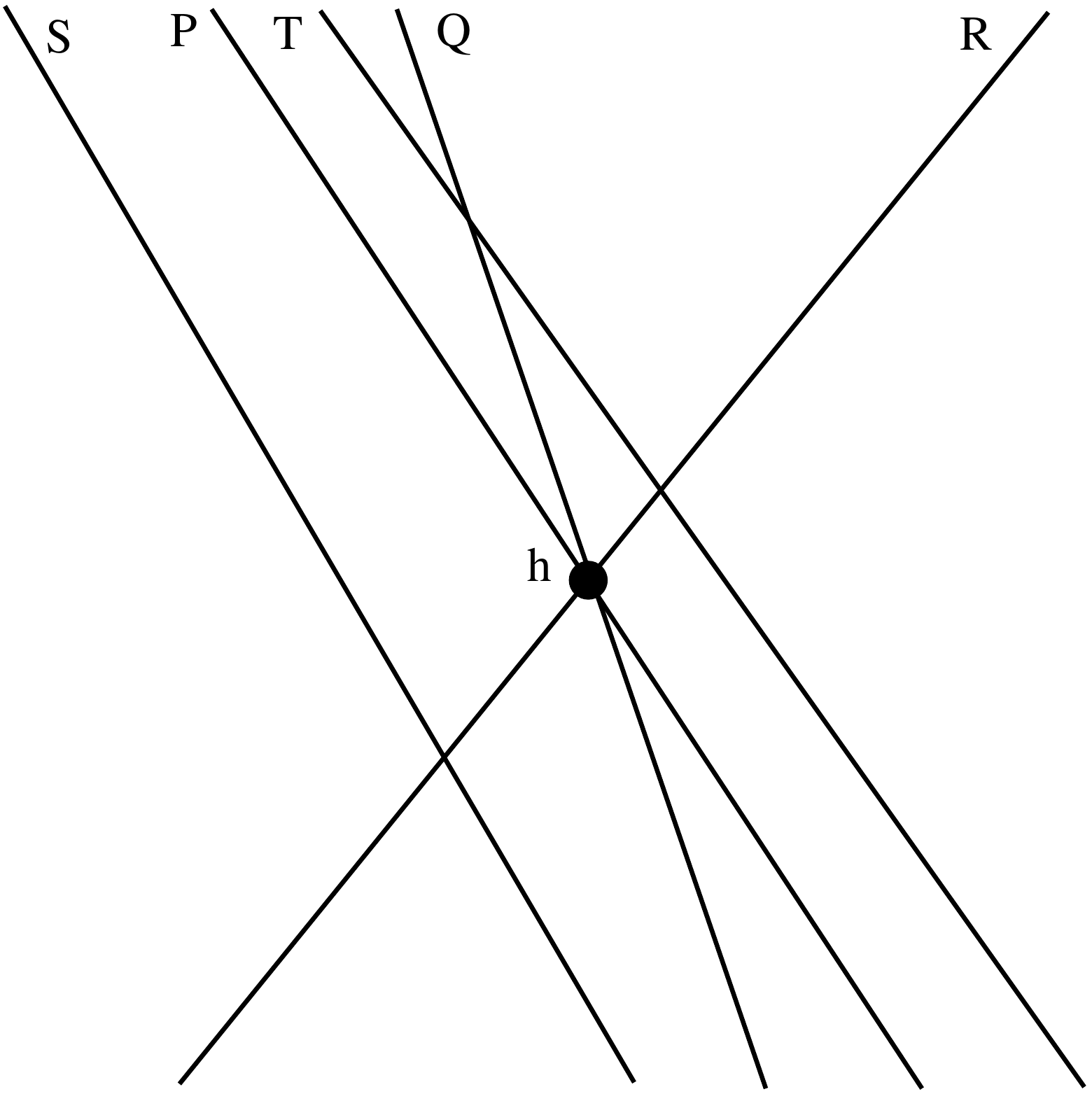}
 \caption{\label{fig:dual}An important property of duality that makes
   it useful for our application is that the duality of non-vertical
   hyperplanes and points preserves relative positions. Dual points
   are labeled by lowercase letters, and dual hyperplanes by capital
   letters.}
\end{center}
\end{figure}

Our goal here is to review how computing part-worth values in choice
based conjoint analysis naturally leads to a \emph{geometrically dual}
formulation of a SVM, see~\cite{giesen-manuscript} for more
details. Assuming that the scale $v$ is linear, then the part-worth
values $v_i(a_{ij})\in\mathbb{R}, a_{ij}\in A_i, i=1,\ldots,n$, should
satisfy constraints of the form
\begin{equation}\label{eq:cbc-cons}
\sum_{i=1}^n v_i(a_i)-v_i(b_i) > 0,
\end{equation}
whenever $a=(a_1,\ldots,a_n)$ was preferred over $b=(b_1,\ldots,b_n)$
by some consumer in a choice experiment. Let $m_i=|A_i|$ and
$m=\sum_{i=1}^n m_i$. Any linear scale $v$ on the domain $A$ is
represented by a vector $\big( v_i(q_{ij})\big)_{i=1,\ldots,n;
 j=1,\ldots,m_i} \in\mathbb{R}^m$, and a choice experiment is defined
by the characteristic vectors $\chi_a\in\{0,1\}^m$, whose $i$'th
component is $1$ if the corresponding parameter level is present in
option $a$, and $0$ otherwise. We can re-write the choice constraints
(\ref{eq:cbc-cons}) as
\[
v^t(\chi_a -\chi_b) > 0, \textrm{ if } a\succeq b \textrm{ in a
 choice experiment,}
\] 
or shorter as $v^tn_{ab} > 0$, where $n_{ab}=\left(\chi_a
 -\chi_b\right)$. A vector $v$ is called \emph{feasible} if it
satisfies all constraints. The set of all feasible vectors is in
general a (not necessarily full-dimensional) double cone whose apex is the
origin. Among all the feasible vectors we want to choose one with good
generalization properties. This can be phrased as a two-class
classification problem as follows: let $H_{ab}$ be the hyperplane
\[
H_{ab} = \{ v\in\mathbb{R}^m\,|\, v^tn_{ab}=0\}
\]
with normal $n_{ab}$, and let
\[
H_{ab}^+ = \{ v\in\mathbb{R}^m\,|\, v^tn_{ab} \geq 0\}
\]
and
\[
H_{ab}^- = \{ v\in\mathbb{R}^m\,|\, v^tn_{ab} \leq 0\}
\]
be the two closed halfspaces bounded by $H_{ab}$. Note that
$H_{ab}^+=H_{ba}^-$. If $a$ was preferred over $b$ in a choice
experiment, then we have a constraint of the form $v\in H_{ab}^+$,
otherwise, if $b$ was preferred over $a$ in a choice experiment, then
we have $v\in H_{ab}^-$. That is, we can assign a label $+$, or $-$,
respectively to the hyperplane $H_{ab}$ depending on the outcome of a
choice experiment for this hyperplane. Since the label attached to the
hyperplane $H_{ba}$ is just the opposite of the label attached to
$H_{ab}$ we can restrict ourselves to one of the two hyperplanes for
every pair $a\neq b\in A$, e.g., by fixing an arbitrary order on the
elements of $A$, and only considering hyperplanes $H_{ab}$, where $a$
comes before $b$ in this order. That is, we are given labelled
hyperplanes as input and are looking for a point in the feasible cone
that can be written as the intersection of the halfspaces $H_{ab}^+$
if $a\succeq b$ in a choice experiment, and $H_{ab}^-$ if $b\succeq a$
in a choice experiment. In standard linear two-class classification
the situation is the other way around: we are given labelled points
and are looking for a hyperplane that separates the points according
to their labels. There are several geometric duality transform know
that map hyperplanes into points and vice versa, see for example~\cite{Edelsbrunner87}, which in principle allow to
transform our problem to compute part-worth values into a standard
two-class classification problem. The duality transform that we
consider here maps non-vertical (labeled) hyperplanes to (labeled)
points and vice versa see Figure~\ref{fig:dual} for an example in
$\mathbb{R}^2$.
Since many of the hyperplanes $H_{ab}$ are vertical, i.e., parallel to
the $m$'th coordinate axis, we augment the hyperplane normals with a
$(m+1)$'th coordinate and set the value of this coordinate to
$\epsilon >0$. This leads to a two-class classification problem that
is parameterized by $\epsilon$. Formulating the SVM for this problem
and taking the limit $\epsilon\rightarrow 0$ leads to the following
QP:
\begin{equation*}
\label{eq:CBC}
\begin{array}{llll}
\mbox{\bf CBC} &\mbox{minimize}_{v}   & \frac{1}{2} \|v\|^2\\
        &\mbox{subject to} & v^tn_{ab} \geq 1, \textrm{ if } a \succeq b \\
        & & \quad \textrm{ in a choice experiment.} \\
\end{array}
\end{equation*}

\paragraph{Soft margin formulation}

On real data we have to deal with contradictory information, i.e.,
observed choices of the form $a\succeq b$ and $b \succeq a$,
especially when assessing preferences on a population, but also
individuals can be inconsistent in their choices. Also with
contradictory information we can proceed as before with the only
difference that after dualizing we work with a soft margin $C$-SVM (\ref{eq:svm}) 
to deal with the contradictions. This leads to the following pQP:
\begin{equation*}
\label{eq:pCBC}
\begin{array}{llll}
\mbox{\bf CBC}$(C)$ &\mbox{minimize}_{v,z}   & \frac{1}{2}\|v\|^2 + C \sum_{j=1}^m \xi_j\\
        &\mbox{subject to} & v^tn_{ab} \geq 1 - \xi_j, \textrm{if }\\
        & & \quad a \succeq b \textrm{ in the }j\textrm{'th choice}\\ 
        & & \quad \textrm{experiment.}\\
& & \xi_j \geq 0,
\end{array}
\end{equation*}
with a non-negative slack variable $\xi_j$ for every choice, i.e., a
constraint $v(a)-v(b) +\xi_j\geq 0, \xi_j \geq 0$ if $a$ was preferred
over $b$ in the $j$'th choice experiment. The slack is penalized in
the objective function by the term $\sum_{j=1}^k \xi_j$ assuming that we
have information from $k$ choice experiments, and $C>0$ is the
standard trade-off parameter between model complexity and quality of
fit on the observed data. The problem {\bf CBC}$(C)$ has already been
suggested by Evgeniou et al.~\cite{EvgeniouBZ05} to compute part-worth
values, but without giving details why it is well suited for that task. A similar resulting formulation is also known as the \emph{ranking SVM} \cite{Joachims:2002p1573}, when the representing features are the present parameter levels in each option.

\section{Experimental  Results}
\label{sec:results}

To test the criss-cross method we provided an experimental proof of
concept implementation. The implementation at its current stage is not
really efficient, but the results for the number of iterations needed
by the criss-cross method along the path are promising. We hope to fully exploit this behavior with a state of the art implementation in the near future.

We tested the criss-cross method on a choice based conjoint analysis
data set that we obtained in a larger user study to measure the
perceived quality for a visualization task~\cite{GiesenMSWZ07}: the
conjoint study had six parameters with $3, 5, 6, 2, 3,$ and $5$
respectively levels. That is, in total this study comprised $24$ levels
for which we estimate the part-worth value. To estimate the part-worth
values we had over participants of our study had to provide answers in
choice experiments. Hence the problem {\bf CBC}$(C)$ leads to problem
{\bf QP}$(\mu)$ whose matrix $Q$ has dimension $(24 + s) \times (24 + s)$, and whose matrix $A$ has dimension $s \times (24 + s)$, when $s$ is the number of
choice experiments. Hence we can essentially control the complexity of the problem by the number of
choice experiments considered.

For $40$ choice experiments exemplary paths need $(580,47,22)$ or
$(580,4,48)$ iterations for the criss-cross method for the first three
bends on a $C$-interval from $[1,6.09 10^{13}]$. This clearly shows that even if a starting point for the solution path may needs some time to be computed ($580$ steps by the criss-cross method, but of course other methods could also be used to compute a starting solution) our described criss-cross method is very effective to continue the path at the bends.

\section{Conclusion}
We have presented a generic solution path algorithm for parameterized
quadratic programs that works for all regularization methods
that result in a single parametric quadratic program, also when the kernel matrix is not
of full rank.

Since the state of the art solution methods in machine learning are moving away from finding exact solutions to faster approximate methods, it would be an interesting further research topic to investigate paths of approximate solutions of parametrized quadratic programs.
Also, it should be further investigated how multi-parametric programming approaches~\cite{Spjtvold:2008p6101} may help to find several parameters simultaneously, such as the regularization parameter, regression tube width, and also kernel parameters~\cite{Wang:2007p106}.

\newpage
\bibliographystyle{abbrv}
\bibliography{../bibliography}

\end{document}